\documentclass[runningheads]{llncs}

\usepackage{graphicx}

\usepackage{tikz}
\usepackage{comment}
\usepackage{amsmath,amssymb} 
\usepackage{color}
\usepackage{algorithm} 
\usepackage{algorithmic} 
\usepackage{bm}
\usepackage{multirow}
\usepackage{booktabs}
\usepackage{orcidlink}

\usepackage[accsupp]{axessibility}  

\begin{document}
\pagestyle{headings}
\mainmatter
\def\ECCVSubNumber{5508}  

\title{Bagging Regional Classification Activation Maps for Weakly Supervised Object Localization}


\titlerunning{Bagging Regional Classification Activation Maps}
%
\author{Lei Zhu\orcidlink{0000-0003-0506-4268} \and
Qian Chen\orcidlink{0000-0003-4475-6582} \and
Lujia Jin\orcidlink{0000-0002-7011-0008} \and 
Yunfei You\orcidlink{0000-0002-9429-2811} \and
Yanye Lu\inst{*}\orcidlink{0000-0002-3063-8051}}
\authorrunning{Lei et al.}
\institute{Institute of Medical Technology, Peking University \and Department of Biomedical Engineering, Peking University \and Institute of Biomedical Engineering, Peking University Shenzhen Graduate School
\email{zhulei@stu.pku.edu.cn, yanye.lu@pku.edu.cn}}

\maketitle

\begin{abstract}

Classification activation map (CAM), utilizing the classification structure to generate pixel-wise localization maps, is a crucial mechanism for weakly supervised object localization (WSOL). However, CAM directly uses the classifier trained on image-level features to locate objects, making it prefers to discern global discriminative factors rather than regional object cues. Thus only the discriminative locations are activated when feeding pixel-level features into this classifier. To solve this issue, this paper elaborates a plug-and-play mechanism called BagCAMs to better project a well-trained classifier for the localization task without refining or re-training the baseline structure. Our BagCAMs adopts a proposed regional localizer generation (RLG) strategy to define a set of regional localizers and then derive them from a well-trained classifier. These regional localizers can be viewed as the base learner that only discerns region-wise object factors for localization tasks, and their results can be effectively weighted by our BagCAMs to form the final localization map. Experiments indicate that adopting our proposed BagCAMs can improve the performance of baseline WSOL methods to a great extent and obtains state-of-the-art performance on three WSOL benchmarks. Code are released at \url{https://github.com/zh460045050/BagCAMs}.
\end{abstract}

\section{Introduction}
\label{sec:intro}

\begin{figure}
\centering
\includegraphics[width=1.00\textwidth]{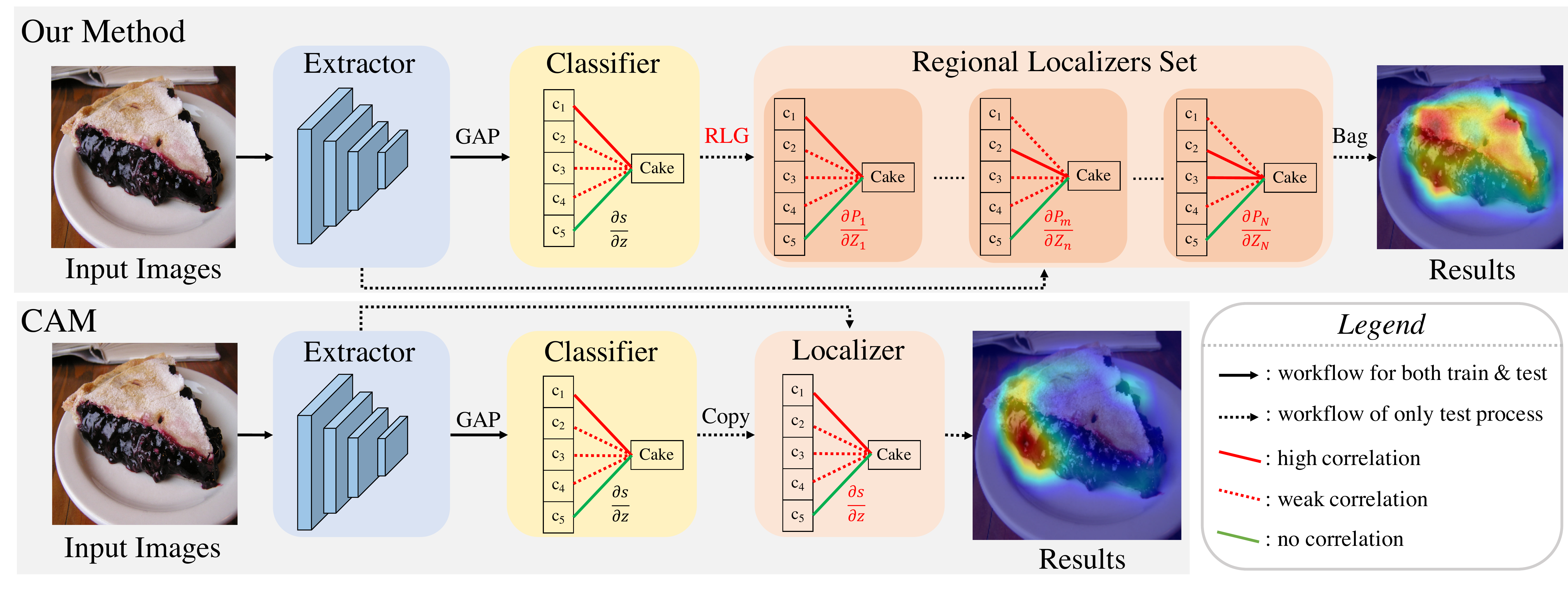}
\caption{Comparison between our BagCAMs and CAM. Our BagCAMs (upper part) derives regional localizers from the classifier with the RLG strategy, while CAM (bottom part) only copies the globally-learned classifier to locate objects.}
\label{fig:intro}
\end{figure}

Weakly supervised learning, using coarse annotations as supervision during model learning, has attracted extensive attention in recent years, especially for localization relevant vision tasks, such as image segmentation~\cite{WSSS2,WSSS1,WSIS} and object detection~\cite{WSDD2,WSDD1}. Typically, weakly supervised object localization (WSOL) releases the requirements of bounding boxes or even the densely-annotated pixel-level localization masks by only learning the localization model with image-level annotations, \textit{i.e.,} the class of images, which effectively saves human resources for the annotation process. The majority of WSOL methods adopt the mechanism of classification activation map (CAM)~\cite{CAM}, utilizing the global average pooling (GAP) to spatially average the pixel-level features into image-level to learn an image classifier with the image-level supervision. Except for generating the classification results, this image classifier also serves as an object localizer that acts on pixel-level features to produce the localization map in the test process.

Though CAM provides an efficient tool for learning a localization model with weak supervision, it directly adopts the classifier as the localizer without considering the difference between them. In detail, the classifier is only learned based on the image-level features, which are spatially aggregated and contain sufficient object features to be discerned. Catching some discriminative factors is enough for the classifier to discern the class of objects. However, the object localizer focuses on discerning the class of all regional positions based on the pixel-level features, where discriminative factors may not be well-aggregated, \textit{i.e.}, insufficient to activate the globally-learned classifier. Thus, the classifier of CAM will only catch the most discriminative parts rather than the whole object locations when directly adopting it to locate objects for pixel-level features.

To solve this issue, a series of methods have been proposed to force the classifier discerning object features more comprehensively, for example, developing augmentation strategies to enrich the global features~\cite{HAS,CUTMIX,CEM}, aligning the feature distribution between image-level and pixel-level~\cite{SPG,DAWSOL}, adopting multi-classifier to synergistically localize the object~\cite{ACOL,DANet,CCAM,CSOA}, or refining the classifier to catch class-agnostic object features~\cite{SEM,SLT}. Though these strategies show some effect, adopting them requires re-training or revising the baseline structure, enhancing the complexity of the training process. Moreover, they still follow CAM to directly adopt the globally-learned classifier as the localizer, indicating that the gap between classifier and localizer remains unresolved.

Unlike the above methods, our work proposes a plug-and-play approach called BagCAMs, which can better project an image-level trained classifier to comply with the requirement of localization tasks. It can easily replace the classifier projection of CAM and be engaged into existing WSOL methods without re-training the network structure. As visualized in Fig.~\ref{fig:intro}, instead of directly adopting the globally-learned classifier, our method focuses on deriving a set of regional localizers from this well-trained classifier. Those regional localizers can discern object-related factors with respect to each spatial position, acting as the base learners of ensemble learning. With those regional localizers, the final localization results can be obtained by integrating their effect. Experiments show that the proposed BagCAMs significantly improves the performance of the baseline methods and achieves state-of-the-art performance on three WSOL benchmarks.

\section{Related Work}

Existing WSOLs can be categorized into multi-stage methods~\cite{UPSP,PSOL,GC,SLT,FAM,ADL} and one-stage methods~\cite{SPG,SEM,CCAM,CEM,I2C,DAWSOL}. The former requires training additional structures upon the classification structure to generate class-agnostic localization results. Our method belongs to the latter, which produces the localization score by projecting the image-classifier back to the pixel-level feature based on CAM, so we just review representative one-stage methods.

To force the classifier to discern some indistinguishable features of objects, Singh \textit{et al.}~\cite{HAS} proposed hide-and-seek (HAS) augmentation that randomly hides the patches of images in the training process. However, hiding patches also causes information loss. Yun \textit{et al.}~\cite{CUTMIX} elaborated a CutMix strategy to solve this issue, which replaces the hidden regions with a patch of another image. Babar~\cite{CAAM} adopts the siamese neural network to align location maps of two images that contain complementary patches of the input. Instead of developing augmentation strategies, some one-stage methods also focus on fusing the localization maps of multiple classifiers to comprehensively catch object parts. Typically, Zhang \textit{et al.}~\cite{ACOL} suggested learning two classifiers to discern features of objects in a complementary way. Kou \textit{et al.}~\cite{CSOA} added an additional classifier to adaptively produce the auxiliary pixel-level mask, which is then utilized by a metric learning loss for supervision. To consider hierarchical cues, Xue \textit{et al.}~\cite{DANet} elaborated the DANet by learning multiple classifiers based on hierarchical features, and Tan \textit{et al.}~\cite{CEM} proposed a pixel-level class selection (PCS) strategy to generalize CAM for hierarchical features. Seunghun \textit{et al.}~\cite{CCAM} fused localization maps of different classes with non-local block~\cite{NL,SNL} to help catch locations that correlated to multiple classes. Compared with them, our BagCAMs generates multiple localizers for each spatial position by degrading a well-trained classifier with efficient post-processing like CAM, rather than re-training the extractor or additional classifiers, increasing the complexity of the training process.

Beyond the community of WSOL, some methods also improved CAM for the visual explanation of convolutional neural networks, \textit{i.e.}, explaining why CNN makes specific decisions. To engage CAM into CNN without the GAP operator, Selvaraju \textit{et al.}~\cite{GradCAM} proposed the GradCAM that summarizes the gradient as the importance of neurons to aggregate feature maps. Aditya \textit{et al.}~\cite{GradCAM++} further improved the GradCAM by elaborating a spatial weighing strategy when summarizing the gradient. Recently, Wang \textit{et al.}~\cite{ScoreCAM} and Desai~\cite{AblationCAM} explored obtaining neuron importance through forward passing to avoid the gradient calculation. Unlike these methods that aim to better activate the discriminative locations, our method focuses on complying CAM mechanism with the purpose of WSOL, activating object locations as many as possible.


\section{Methodology}
\label{sec:method}

This section first formally overviews our proposed method that localizes objects with a series of regional localizers. Then, the regional localizer generation (RLG) strategy is illustrated, helping generate these regional localizers for the localization task. Finally, the BagCAMs is proposed to derive these localizers from a well-trained image classifier and produce the final localization map.

\subsection{Problem Definition}

Given an input image represented by $\bm{X} \in \bbbr^{3 \times N^{I}}$, WSOL aims to approximate the localization map $\bm{Y} \in \bbbr^{K \times N^{I}}$ by a localization model learned only with the image-level classification mask $\bm{y} \in \bbbr^{K \times 1}$, where $K$ and $N^{I}$ are the numbers of classes of interest and pixels, respectively. To learn the localization model with $\bm{y}$, a backbone network, \textit{i.e.}, ResNet~\cite{RESNET} or InceptionV3~\cite{INCEPTION}, is firstly adopted as the feature extractor $e(\cdot)$ to extract pixel-level features $\bm{Z} = e(\bm{X}) \in \bbbr^{C \times N}$, where $C$ is the channels of the features with the spatial resolution $N$. These pixel-level features are fed into the GAP layer to generate the image-level feature $\bm{z} \in \bbbr^{C \times 1}$. Finally, the classifier $c(\cdot)$ implemented as the fully-connected layer with weight $\mathbf{W} \in \bbbr^{K \times C}$ is acted on the image-level feature to generate the classification result $\bm{s}$:
\begin{equation}
\bm{s}_{k} = c(\bm{z})_{k} = (\mathbf{W}\bm{z})_{k} = \sum_{c} \mathbf{W}_{k, c} \bm{z}_{c}~~,
\label{eq:CAM_cls}
\end{equation}
\noindent where $k$ and $c$ are the index of class and channel, respectively. This classification score $\bm{s}$ is supervised by the cross-entropy $\mathcal{L}_{ce}(\bm{y}, \bm{s})$ to learn the extractor $e(\cdot)$ and the classifier $c(\cdot)$ in the training process.

In the test process, except for generating the classification score $\bm{s}$, CAM-based methods also utilize the classifier $c(\cdot)$ as a localizer $f(\cdot)$ that acts onto the pixel-level features $\bm{Z}$ to obtain the localization maps $\bm{P} \in \bbbr^{K \times N}$:
\begin{equation}
\bm{P}_{k, i} = f(\bm{Z})_{k, i} = c(\bm{Z}_{:, i})_{k} = \sum_{c} \mathbf{W}_{k, c} \bm{Z}_{c, i}~~.
\label{eq:CAM}
\end{equation}

As discussed in Sec.~\ref{sec:intro}, the classifier $c(\cdot)$ is only learned based on the image-level feature $\bm{z}$, which aggregates the object features on all the positions of $\bm{Z}$. This makes the classifier $c(\cdot)$ only discern the most discriminative feature rather than all features that are correlated to the objects. When directly projecting the classifier $c(\cdot)$ as the localizer $f(\cdot)$ that acts on the pixel-level features, some indistinguishable parts, \textit{i.e.}, the body of animals, will not be activated on the output localization maps $\bm{P}$. Thus, our method adopts the proposed RLG strategy to generate a base localizer set $\mathcal{F} = \{ f_{1}, f_{2}, ..., f_{n} \}$ to comprehensively discern the feature of objects. Then, the proposed BagCAMs can implement the base localizer set $\mathcal{F}$ based on the image-classifier $c(\cdot)$ and generate a series of localization maps $\mathcal{P} = \{ \bm{P}_{1}, \bm{P}_{2}, ..., \bm{P}_{n} \}$. Finally, these maps are integrated with co-efficient $\{ \lambda_{1}, \lambda_{2}, ..., \lambda_{n} \}$ to form the final localization map $\bm{P}^{*}$ that determines $\bm{Y}$:
\begin{equation}
\bm{P}^{*} = \sum_{n} \lambda_{n} f_{n}(\bm{Z})~~.
\end{equation}

\subsection{Regional Localizers Generation Strategy}

The proposed RLG strategy utilizes localization scores and pixel-level feature maps to generate a set of regional localizers, which focuses more on the regional features rather than only discerning the global features as the classifier of the classification task. To better illustrate the proposed RLG strategy, we firstly design the regional localizer inspired by the property of an image classifier. In detail, by differentiating from Eq.~\ref{eq:CAM_cls}, the weight $\mathbf{W}$ of the global classifier $c(\cdot)$ can be reformulated~\cite{CEM}:
\begin{equation}
\mathbf{W} = \frac{\partial c(\bm{z})}{\partial \bm{z}} = (\frac{\partial \bm{s}}{\partial \bm{z}})^{\top}~~.
\end{equation} 
\noindent Taking it into the Eq.~\ref{eq:CAM_cls}, a equivalency of the classifier $c(\cdot)$ can be obtained~\cite{CEM}:
\begin{equation}
c(\bm{z}) = \mathbf{W} \bm{z} = (\frac{\partial \bm{s}}{\partial \bm{z}})^{\top} \bm{z}~~.
\label{eq:sz}
\end{equation}

\noindent Eq.~\ref{eq:sz} indicates that an image classifier $c(\cdot)$ can be represented by the transposition of the partial derivative between the image classification score $\bm{s}$ and the image feature $\bm{z}$~\cite{CEM}. Analogizing this property to the localization task, the regional localizer can be simulated with the following definition.
\begin{definition}
Assuming $f(\cdot)$ is a localizer that generates the classification score $\bm{p}$ on a specific spatial location based on the pixel-level features $\bm{Z} \in \bbbr^{C \times N}$, \textit{i.e.}, $\bm{p} = f(\bm{Z})$, the localizer $f(\cdot)$ can be simulated by a function set $\mathcal{F}$ that contains the partial derivative between this regional classification score $\bm{p}$ and each regional position of pixel-level features $\bm{Z}$:
\begin{equation}
\mathcal{F} = \{ f_{1}, ..., f_{n}, ..., f_{N} \} = \{ (\frac{\partial \bm{p}}{\partial \bm{Z}_{:, 1}})^{\top}, ..., (\frac{\partial \bm{p}}{\partial \bm{Z}_{:, n}})^{\top}, ..., (\frac{\partial \bm{p}}{\partial \bm{Z}_{:, N}})^{\top} \}~~,
\end{equation}
\noindent where $f_{n}(\cdot) = (\frac{\partial \bm{p}}{\partial \bm{Z}_{:, i}})^{\top} (\cdot)$ is the regional localizer that catches the relation between regional score $\bm{p}$ and the pixel-level feature of a specific regional position $\bm{Z}_{:, i}$. 
\label{definition}
\end{definition}

\begin{figure}
\centering
\includegraphics[width=1.00\textwidth]{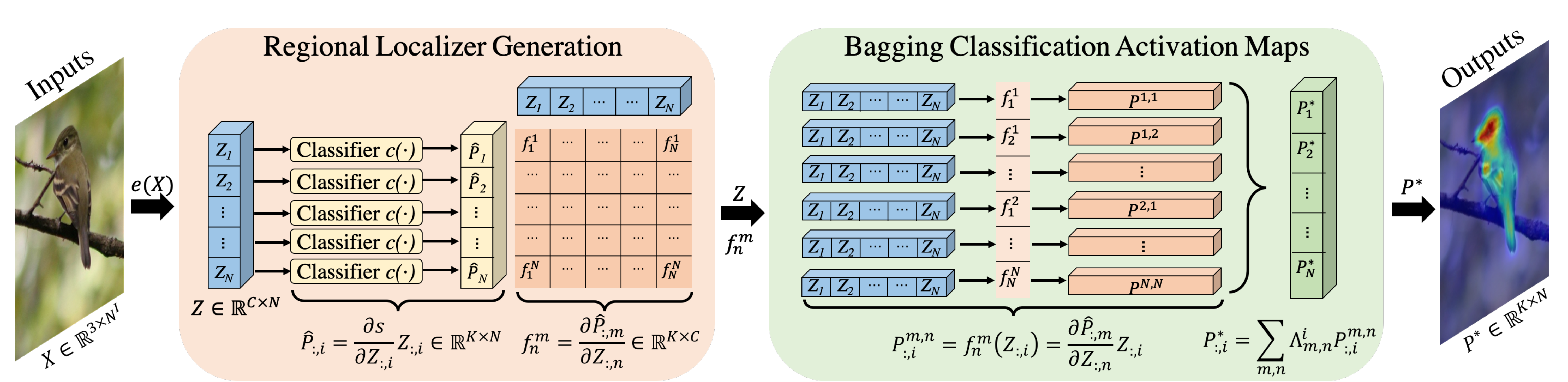}
\caption{Workflow of our method, where the RLG strategy (orange) generates a set of classifiers and the BagCAMs (green) weights their effect to produce localization maps.}
\label{fig:structure}
\end{figure}

Based on Definition~\ref{definition}, each row vector $\bm{P}_{:, i}$ of a given localization map $\bm{P} \in \bbbr^{K \times N}$ can be viewed as a regional classification score $\bm{p}$ that defines $N$ regional localizers based on the pixel-level feature $\bm{Z}$. Thus, as indicated in Fig.~\ref{fig:structure}, our RLG strategy (noted by orange) can simulate $N*N$ regional localizers based on the correlation between each vector pair of $\bm{P}$ and $\bm{Z}$:
\begin{equation}
f^{m}_{n}(\bm{x}) = (\frac{\partial \bm{P}_{:, m}}{\partial \bm{Z}_{:, n}})^{\top} (\bm{x}) 
~\longrightarrow~ f^{m}_{n}(\bm{x})_{k} = \sum_{c} \frac{\partial \bm{P}_{k, m}}{\partial \bm{Z}_{c, n}} \bm{x}_{c} ~~,
\label{eq:base_learner}
\end{equation}
\noindent where $f^{m}_{n}(\cdot)_{k}$ represents the regional localizer of class $k$ and $\bm{x} \in \bbbr^{C \times 1}$ is a variable that represents a feature vector. With this extension, a localizer set $\mathcal{F}^{*}$ that contains $N*N$ regional localizer can be defined based on $\bm{P}$ and $\bm{Z}$, \textit{i.e.,} $\mathcal{F}^{*} = \{ f^{1}_{1}, ...,  f^{m}_{n}, ..., f^{N}_{N} \}$. Compared with the global classifier $(\frac{\partial \bm{s}}{\partial \bm{z}})^{\top}$ used by CAM, our regional localizer set $\mathcal{F}^{*}$ contains sufficient localizers that catch the regional correlation between scores and features on each position, which helps comprehensively discern features of the objects.
\subsection{Bagging Regional Classification Activation Maps}
\label{sec:genrealize}
The proposed RLG strategy provides an efficient mechanism to generate a localizer set $\mathcal{F}^{*}$ based on the localization map $\bm{P}$. When implementing $\bm{P}$ as a coarse localization map $\hat{\bm{P}} \in \bbbr^{K \times N}$, those regional localizers $f^{m}_{n}$ can be viewed as the base learners that can be integrated as a strong learner to locate objects. For this purpose, our BagCAMs is proposed as shown in Fig.~\ref{fig:structure} (noted by green), which generates the base localizers based on a coarse localization map $\hat{\bm{P}}$ and then weights their localization results as the final localization score:
\begin{equation}
\bm{P}^{*}_{k, i} = \sum_{m} \sum_{n} \mathbf{\Lambda}^{i}_{m,n} f^{m}_{n}(\bm{Z}_{:, i})_{k} = \sum_{m} \sum_{n} \mathbf{\Lambda}^{i}_{m,n} \sum_{c} \frac{\partial \hat{\bm{P}}_{k, m}}{\partial \bm{Z}_{c, n}} \bm{Z}_{c, i} ~~, 
\label{eq:bagcam}
\end{equation}
\noindent where $\bm{P}^{*}$ is the localization map of our proposed BagCAMs whose element $\bm{P}^{*}_{k, i}$ represents the score on class $k$ at position $i$. $\mathbf{\Lambda}^{i}$ is a matrix, and its element $\mathbf{\Lambda}^{i}_{m,n}$ means the co-efficient of regional localizer $f^{m}_{n}$ at position $i$. In detail, PCS strategy~\cite{CEM} is adopted to initialize the coarse localization map $\bm{\hat{P}}_{k, m}$ to pursue the convenience of calculation and performance on intermediate feature maps:
\begin{equation}
\bm{\hat{P}}_{k, m} = \sum_{c} \frac{\partial \bm{s}_{k}}{\partial \bm{Z}_{c, m}}{\bm{Z}_{c, m}}~~.
\label{eq:setting}
\end{equation}

\begin{table*}[!htbp]
\centering
\caption{Summary of degrading the proposed BagCAMs into other methods}
\begin{tabular}{c|c|c|c}
\toprule
 & Initial Score $\hat{\bm{P}}_{k, m}$ & Co-efficient Matrix $\mathbf{\Lambda}^{i}$ & Localization Score $\bm{P}^{*}_{k, i}$ \\
\hline
CAM & $\bm{s}_{k}$ & $\mathbf{\Lambda}^{i}=\frac{1}{N}\mathbf{I}$ & $\sum_{c}\frac{\partial \bm{s}_{k}}{\partial \bm{z}_{c}} \bm{Z}_{c, i}$ \\
\hline
GradCAM & $\bm{s}_{k}$ & $\mathbf{\Lambda}^{i}=\frac{1}{N}\mathbf{I}$ & $\frac{1}{N}\sum_{n,c}\frac{\partial \bm{s}_{k}}{\partial \bm{Z}_{c,n}} \bm{Z}_{c, i}$ \\
\hline
GradCAM++ & $\bm{s}_{k}$ & $\mathbf{\Lambda}^{i}=diag(\mathbf{\alpha})$  & $\sum_{n,c} \mathbf{\alpha}_{m} \frac{\partial \bm{s}_{k}}{\partial \bm{Z}_{c,n}} \bm{Z}_{c, i}$ \\
\hline
PCS & $\bm{s}_{k}$ & $\mathbf{\Lambda}^{i}_{m, n}=\left\{
\begin{array}{rcl}
1    &  ,~~    & {i = n}\\
0     &   ,~~   & {i \neq n}
\end{array} \right.$ & $\sum_{c}\frac{\partial \bm{s}_{k}}{\partial \bm{Z}_{c,i}} \bm{Z}_{c, i}$ \\
\hline
\textbf{Ours} & $\sum_{c} \frac{\partial \bm{s}_{k}}{\partial \bm{Z}_{c, m}}{\bm{Z}_{c, m}}$ & $\mathbf{\Lambda}^{i}_{m, n}=\left\{
\begin{array}{rcl}
1    &  ,~~    & {i = n}\\
0     &   ,~~   & {i \neq n}
\end{array} \right.$ & $\sum_{m,c_{2}} \frac{\partial (\sum_{c_{1}} \frac{\partial \bm{s}_{k}}{\partial \bm{Z}_{c_{1}, m}} \bm{Z}_{c_{1}, m})}{\partial \bm{Z}_{c_{2}, i}} \bm{Z}_{c_{2}, i}$ \\
\bottomrule
\end{tabular}
\label{tab:relation}
\end{table*}

\noindent With this initialization coarse localization map $\bm{\hat{P}}_{k, m}$ and defining $\bar{\bm{s}}= log(\bm{s})$, the formulation of our base localizer generated by our RLG derivates into the following, whose proof are given in Appendix B:
\begin{equation}
\begin{split}
f^{m}_{n}(\bm{x})_{k} = \sum_{c_{1}} \bm{s}_{k}(1 + \frac{\partial \bar{\bm{s}}_{k}}{\partial \bm{Z}_{c_{1}, m}}\bm{Z}_{c_{1}, m})  \sum_{c_2} (\frac{\partial \bar{\bm{s}}_{k}}{\partial \bm{Z}_{c_{2}, n}} \bm{x}_{c_{2}}) ~~.
\label{eq:bagcem_f}
\end{split}
\end{equation}

\noindent As for the weight matrix $\mathbf{\Lambda}^{i}$, the grouping strategy of PCS~\cite{CEM} is also adopted for consistency, assuming $(\frac{\partial \bm{\bm{p}}}{\partial \bm{Z}_{:, i}})^{\top}$ is the localizer specifically for the position $i$:
\begin{equation}
\mathbf{\Lambda}^{i}_{m, n}=\left\{
\begin{array}{rcl}
1    &  ,~~    & {i = n}\\
0     &   ,~~   & {i \neq n}
\end{array} \right. ~~.
\label{eq:weight}
\end{equation}
\noindent This setting assigns the $N*N$ regional localizers into $N$ groups, each applied specifically to position $i$. Note that $\mathbf{\Lambda}^{i}$ can also be implemented with other mechanisms, for example, spatial average~\cite{GradCAM} or spatial attention~\cite{GradCAM++}, but we find the grouping strategy performs the best due to lesser noise. Finally, taking Eq.~\ref{eq:bagcem_f} and Eq.~\ref{eq:weight} into Eq.~\ref{eq:bagcam}, an executable formulation of BagCAMs is obtained:
\begin{equation}
\bm{P}^{*}_{k, i}  = \sum_{m} \sum_{c_{1}} \bm{s}_{k}(1+\frac{\partial \bar{\bm{s}}_{k}}{ \partial \bm{Z}_{c_{1}, m}}\bm{Z}_{c_{1}, m}) (\sum_{c_{2}} \frac{\partial \bar{\bm{s}}_{k}}{ \partial \bm{Z}_{c_{2}, i}} \bm{Z}_{c_{2}, i})~~.
\label{eq:bagcem}
\end{equation}

As indicated in Eq.~\ref{eq:bagcem}, the computation of our BagCAMs only relies on the gradients $\frac{\partial \bar{\bm{s}}}{\partial \bm{Z}}$, which can be calculated by backward propagating gradients on the logarithm of the classification score $\bm{s}$. Thus, our BagCAMs can be projected onto the intermediate layer of CNN and retain similar computation complexity as gradient-based CAM mechanisms~\cite{GradCAM,GradCAM++,CEM}. Moreover, Table~\ref{tab:relation} also shows PCS~\cite{CEM} and other CAM mechanisms~\cite{CAM,GradCAM,GradCAM++} can also be generalized by our BagCAMs with the assumption that the initial localization result of each position $i$ are all equal to $\bm{s}_{k}$, \textit{i.e.}, $\forall \bm{\hat{P}}_{k, m}=\bm{s}_{k}$. However, this assumption is obviously invalid for the localization task because the background locations of the image should not have the same score as the object locations. Compared with them, our BagCAMs generates a specific initial score $\bm{\hat{P}}_{k, m} \in \mathbb{R}^{K \times N}$ for each position to obtain more valid base localizers to generate high-quality localization maps, rather than defining the localizer only based on the global score $\bm{s} \in \mathbb{R}^{K \times 1}$. This makes our BagCAMs perform much better than these mechanisms when engaged into WSOL.

The proposed BagCAMs can easily replace CAM step of WSOL methods to generate the localization maps. Algorithm~\ref{alg1} and Fig.~\ref{fig:structure} show the workflow of localizing objects for an input image $\bm{X}$ based on a trained WSOL model that contains a feature extractor $e(\cdot)$ and a classifier $c(\cdot)$. Specifically, the input image $\bm{X}$ is firstly fed into the feature extractor $e(\cdot)$ to generate the pixel-level feature $\bm{Z} = e(\bm{X})$. Then, $\bm{Z}$ is aggregated into image-level feature $\bm{z}$, which is fed into the classifier to produce the classification score $\bm{s}$ determining the object class $k=\arg\max (\bm{s})$. Next, backward propagation is adopted for $\bar{\bm{s}}_{k}$ to calculate $\frac{\partial \bar{\bm{s}}_{k}}{\partial \bm{Z}}$ that is crucial for defining the base localizer. Finally, the localization map $\bm{Y}$ is obtained by weighing the localization scores of base localizers as in Eq.~\ref{eq:bagcem}.
\begin{algorithm}
\caption{Workflow of BagCAMs for a Given WSOL Model} 
\label{alg1} 
\begin{algorithmic}[1] 
\REQUIRE Input image $\bm{X}$, Classifier $c(\cdot)$, Extractor $e(\cdot)$.
\STATE Calculating pixel-level feature $\bm{Z}$ of input image $\bm{X}$ with extractor $e(\bm{X})$.
\STATE Obtaining image-level feature $\bm{z}$ with GAP or other aggregation mechanisms.
\STATE Generating image classification score $\bm{s}$ with classifier $c(\bm{z})$.
\STATE Calculating classification results $k = \arg\max(\bm{s})$.
\STATE Backward propagating $\bar{\bm{s}} = \log(\bm{s})_{k}$ to obtain the gradient $\frac{\partial \bar{\bm{s}}_{k}}{\partial \bm{Z}}$.
\STATE Generating BagCAMs localization map $\bm{P}^{*}_{k, :}$ by Eq.~\ref{eq:bagcem} and upsampling it as $\bm{Y}$.
\ENSURE  Localization Score $\bm{Y}$, Classification Score $\bm{s}$.
\end{algorithmic} 
\end{algorithm}

\section{Experiments}
This section first introduces the setting of experiments. Then, results of our BagCAMs are shown to compare with SOTA methods on three datasets. Finally, we  investigate different settings of our BagCAMs to further reflect its validity.

\subsection{Settings}

The proposed BagCAMs can be engaged into a well-trained WSOL model by simply replacing CAM in the test process. Thus, we reproduced five WSOL methods as the baseline methods to train them with their optimal settings, including CAM~\cite{CAM}, HAS~\cite{HAS}, CutMix~\cite{CUTMIX}, ADL~\cite{ADL}, and DAOL~\cite{DAWSOL}. In detail, the ResNet-50, removing the down-sample layer of $Res_4$, was used as the feature extractor. When using InceptionV3 as the extractor, we follow existing works~\cite{UPSP,SPG,ACOL,CEM} that add two additional layers at the end of the original structure. The classifier is implemented as a fully-connected layer, whose outputs are supervised by the cross entropy based on the image-level annotation in the training process. Except for the method-specific strategy~\cite{HAS,CUTMIX,ADL},  the random resize with size $256 \times 256$ and random horizontal flip crop with size $224 \times 224$ were adopted as the augmentation. SGD with weight decay $10^{-4}$ and momentum $0.9$ was set as the optimizer. Note that the learning rate and the method-specific hyper-parameters for all datasets were adopted as the released optimal settings~\cite{EVAL,DAWSOL}. In the test process, our BagCAMs replaced the CAM step of these methods to project the learned classifier as the localizer based on features outputted by $Res_3$ of the ResNet ($Mix_{6e}$ for the InceptionV3). All experiments were implemented with Pytorch toolbox~\cite{PYTORCH} on an Intel Core i9 CPU and an NVIDIA RTX 3090 GPU.

Three standard benchmarks were utilized to evaluate our methods:
\begin{itemize}
\item {\textbf{CUB-200 dataset}}~\cite{CUB} contains 11,788 images that are fine-grained annotated for 200 classes of birds. We follow the official training/test split to use 5,944 images as the training set that only utilizes image-level annotation to supervise WSOL methods. Other 5,794 images, given additional bounding boxes and pixel-level masks, serve as the test set to evaluate the performance. 
\item {\textbf{ILSVRC dataset}}~\cite{IMAGENET} contains 1.3 million images that include 1000 classes of objects. Among them, 50,000 images, whose bounding boxes annotation is provided, are adopted as the test set to report the localization performance.
\item {\textbf{OpenImages dataset}}~\cite{OPENIMAGE,EVAL} contains 37,319 images of 100 classes, where 29,819 images serve as the training set. Following the split released by Junsuk~\cite{EVAL}, the rest 7,500 images, annotated by pixel-level localization mask, are divided into the validation set (2,500 images) and test set (5,000 images).
\end{itemize}

\noindent Note that our BagCAMs does not contain any hyper-parameters, thus only the test images of these dataset are utilized for comparison. The Top-1 localization accuracy (T-Loc)~\cite{HAS}, ground-truth known localization accuracy (G-Loc)~\cite{HAS}, and the recently proposed MaxBoxAccV2~\cite{EVAL} (B-Loc) were adopted to evaluate the performance based on bounding box annotations. As for pixel-level localization masks, the peak intersection over union (pIOU)~\cite{SEM} and the pixel average precision (PxAP)~\cite{EVAL} were calculated as the metrics. 

\begin{table*}[!htp]
\caption{Comparison with SOTA methods with ResNet50 (\textbf{border} means the best).}
\centering	
\setlength{\tabcolsep}{2.4pt}
\begin{tabular}{c|ccccc|ccc|cc}
\toprule
~  & \multicolumn{5}{c|}{CUB-200} & \multicolumn{3}{c|}{ILSVRC} & \multicolumn{2}{c}{OpenImages}\\
 & T-Loc & G-Loc & B-Loc & pIoU & PxAP & T-Loc & G-Loc & B-Loc & pIoU & PxAP\\
\hline
DGL\cite{CEM}
& 60.82 & 76.65 & - & - & -
& 53.41 & 66.52 & - 
& - & -
\\
CAAM\cite{CAAM}
& 64.70 & 77.35 & - & - & -
& 52.36 & 67.89 & - 
& - & -
\\
DANet\cite{DANet}
& 61.10 & - & - & - & -
& - & - & -
& - & -
\\
ICLCA\cite{ICLCA}
& 56.10 & 72.79 & 63.20 & - & -
& 48.40 & 67.62 & 65.15
& - & -
 \\
PAS\cite{PAS}
& 59.53 & 77.58 & 66.38 & - & -
& 49.42 & 62.20 & 64.72 
& - & 60.90
\\
IVR\cite{IVR}
& - & - & 71.23 & - & -
& - & - & 65.57
& - & 58.90
\\
\underline{PSOL}\cite{PSOL}
& 70.68 & - & - & - & -
& 53.98 & 65.54 & -
& - & -
 \\
\underline{SEM}\cite{SEM}
& - & - & - & - & -
& 53.84 & 67.00 & -
& - & -
 \\
\underline{FAM}\cite{FAM}
& \textbf{73.74} & 85.73 & - & - & -
& \textbf{54.46} & 64.56 & -
& - & -
\\
\hline
CAM\cite{CAM} 
& 55.31 & 66.06 & 59.21 & 46.70 & 65.94
& 49.93 & 67.30 & 62.69
& 43.13 & 57.88
\\
+\textbf{Ours}
& 70.89 & 87.44 & 76.22 & 64.40 & 84.38
& 52.14 & 70.78 & 69.13
& 47.92 & 62.52 \\
\hline
HAS\cite{HAS}
& 54.48 & 72.55 & 66.25 & 51.00 & 71.87
& 50.80 & 66.91 & 64.67
& 42.28 & 55.83
\\
+\textbf{Ours} 
& 65.93 & 89.65 & 84.45 & 70.24 & 88.94
& 53.32 & 70.67 & 69.17
& 47.71 & 62.45
\\
\hline
CutMix\cite{CUTMIX}  
& 56.27 & 64.13 & 59.08 & 44.21 & 65.23
& 50.17 & 65.84 & 63.73
& 42.85 & 57.97
\\
+\textbf{Ours} 
& 72.96 & 87.44 & 79.67 & 64.93 & 85.36
& 53.02 & 69.92 & 68.53
& 46.67 & 60.16
 \\
\hline
ADL\cite{ADL} 
& 52.13 & 66.75& 59.31& 45.40& 59.49
& 50.40 & 66.88 & 64.50
& 42.29 & 56.21
\\
+\textbf{Ours} 
& 64.41 & 86.06 & 74.48 & 60.46 & 81.07
& 53.05 & 70.51 & 68.97
& 47.04 & 61.76
\\
\hline
DAOL\cite{DAWSOL}
& 62.40 & 81.83 & 69.87 & 56.18 & 74.70
& 43.26 & 70.27 & 68.23
& 49.68 & 65.42
\\
+\textbf{Ours} 
& 69.67 & \textbf{94.01} & \textbf{84.88} & \textbf{74.51} & \textbf{90.38}
& 44.24 & \textbf{72.08} & \textbf{69.97}
& \textbf{52.17} & \textbf{67.68}
\\
\bottomrule
\end{tabular}
\label{tab:results}
\end{table*}

\begin{table*}[!htp]
\caption{Comparison with SOTA methods with InceptionV3 (\textbf{border} means the best).}
\centering	
\setlength{\tabcolsep}{2.4pt}
\begin{tabular}{c|ccccc|ccc|cc}
\toprule
~  & \multicolumn{5}{c|}{CUB-200} & \multicolumn{3}{c|}{ILSVRC} & \multicolumn{2}{c}{OpenImages}\\
Method & T-Loc & G-Loc & B-Loc & pIoU & PxAP & T-Loc & G-Loc & B-Loc & pIoU & PxAP\\
\hline
DGL\cite{CEM}
& 50.50 & 67.64 & - & - & -
& 52.23 & 68.08 & -
& - & -
\\
SPG\cite{SPG}
& 56.64 & - & - & - & -
& 49.60 & 64.69 & -
& - & -
\\
I$^2$C\cite{I2C}
& 55.99 & 72.60 & - & - & -
& 53.11 & 68.50 & -
& - & -
\\
ICLCA\cite{ICLCA}
& 56.10 & 67.93 & - & - & -
& 49.30 & 65.21 & -
& - & -
\\
PAS\cite{PAS}
& 69.96 & 73.65 & - & - & -
& 50.56 & 64.44 & -
& - & 63.30
\\
IVR\cite{IVR}
& - & - & 61.74 & - & -
& - & - & 66.04
& - & 64.08
\\
UPSP\cite{UPSP}
& 53.38 & 72.14 & - & - & -
& 52.73 & 68.33 & -
& - & -
 \\
\underline{PSOL}\cite{PSOL}
& 65.51 & - & - & - & -
& 54.82 & 65.21 & -
& - & -
 \\
\underline{GCNet}\cite{GC}
& - & - & - & - & -
& 49.06 & - & -
& - & -
 \\
\underline{FAM}\cite{FAM}
& \textbf{70.67} & 87.25 & - & - & -
& \textbf{55.24} & 68.62 & -
& - & -
 \\
\hline
CAM\cite{CAM} 
& 48.96 & 63.44 & 57.14 & 49.28 & 70.95
& 50.75 & 66.16 & 63.61
& 47.51 & 63.31
\\
+\textbf{Ours}
& 54.75 & 74.75 & 65.65 & 60.34 & 81.49
& 52.22 & 68.84 & 66.46
& 49.98 & 65.91 \\
\hline
HAS\cite{HAS}
& 52.68 & 70.89& 62.39& 52.78& 74.07
& 51.00 & 66.99 & 64.26
& 42.87 & 59.50
\\
+\textbf{Ours} 
& 57.93 & 79.44& 69.65& 61.75& 83.03
& 52.22 & 69.20  & 66.89
& 48.44 & 64.37
\\
\hline
CutMix\cite{CUTMIX}
& 51.86 & 66.62& 59.44& 51.40& 74.19
& 50.72 & 66.96 & 64.44
& 46.30 & 62.12
\\
+\textbf{Ours} 
& 58.48 & 79.58& 68.09& \textbf{62.44}& \textbf{83.15}
& 52.60 & 70.57 & \textbf{68.04}
& 49.28 & 65.23
\\
\hline
ADL\cite{ADL}
& 49.10 & 62.62& 57.01& 49.72& 70.06
& 50.20 & 66.30 & 63.66
& 47.03 & 63.42
\\
+\textbf{Ours} 
& 54.75 & 74.34& 64.87& 60.09& 81.41
& 51.63 & 68.81 & 66.42
& 49.22 & 65.31
\\
\hline
DAOL\cite{DAWSOL}
& 56.29 & 80.03 & 68.01 & 51.80 & 71.03
& 52.70 & 69.11 & 64.75
& 48.01 & 64.46
\\
+\textbf{Ours} 
& 60.07 & \textbf{89.78} & \textbf{76.94} & 58.05 & 72.97
& 53.87 & \textbf{71.02} & 66.93
& \textbf{50.79} & \textbf{66.89}
\\
\bottomrule
\end{tabular}
\label{tab:results_inc}
\end{table*}

\subsection{Comparison with state-of-the-arts}

\begin{figure}
\centering
\includegraphics[width=1.00\textwidth]{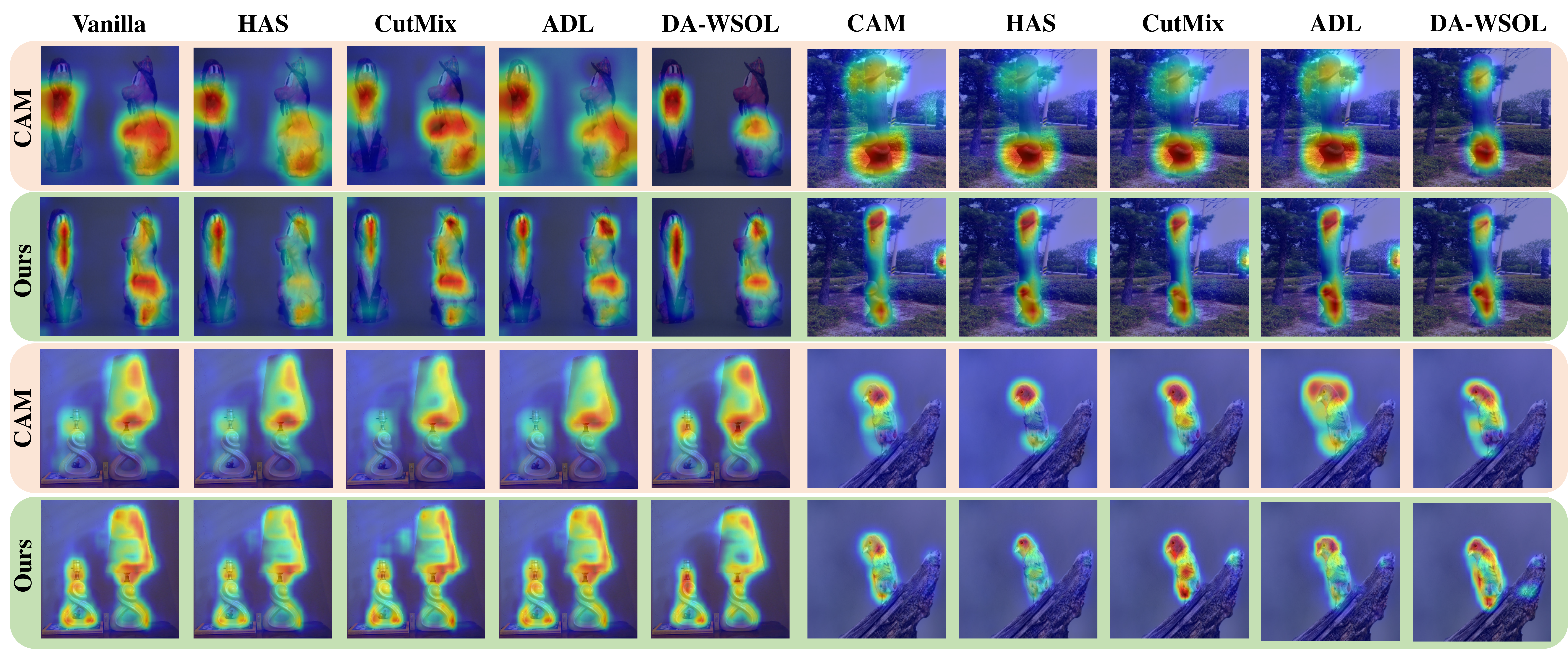}
\caption{Visualization on replacing CAM into BagCAMs for different WSOL methods.}
\label{fig:results}
\end{figure}

Table~\ref{tab:results} illustrates the results of SOTA methods and our BagCAMs on the three standard WSOL benchmarks. It shows that adopting our proposed BagCAMs improves the performance of baseline methods to a great extent, especially on the CUB-200 dataset. This is because the CUB-200 dataset is a fine-graining dataset that only contains birds, making the classifier more likely to catch discriminative parts rather than the common parts of birds. As discussed in Sec.~\ref{sec:method}, this situation basically causes unsatisfactory performance when directly using the classifier to localize objects as CAM. By adopting our BagCAMs to project the classifier into a set of regional localizers, the regional factors of the class of bird can be better concerned, improving nearly 21.38\% in G-Loc than the baseline method. Additionally, when more finely evaluated by the pixel-level mask, the improvements of our method are still remarkable, achieving 64.40\% pIoU and 84.38\% PxAP, which are 17.70\% and 18.44\% higher than CAM, respectively. As for the larger scale dataset ILSVRC, directly replacing CAM into our BagCAMs in the test process also achieves 3.48\% higher performance in G-Loc metric, \textit{i.e.}, correcting the localization of nearly 1,740 images without any fine-tuning process or structure modification. In addition, even using the most recently proposed DAOL~\cite{DAWSOL} that achieves the SOTA performance on the OpenImages dataset, adopting our method can still obviously improve its performance about $2.49\%$ and $2.26\%$, respectively in the pIOU and PxAP.

Except for the five reproduced methods, the other nine SOTA methods were also used for comparison in Table~\ref{tab:results}, whose scores are cited from corresponding papers. Our BagCAMs outperforms SOTA methods for nearly all metrics on all three datasets even though engaged in the vanilla WSOL structure, \textit{i.e.}, ``CAM + Ours". Only the T-Loc metric of our BagCAMs is lower than the methods that generate class-agnostic localization results and adopt addition stages for classification (noted by \underline{underline style})~\cite{FAM,SEM,PSOL}. This is because our BagCAMs is only adopted in the test process to enhance the localization results, and our classification accuracy is directly determined by the baseline WSOL methods. Moreover, Table~\ref{tab:results_inc} also shows the comparison of using InceptionV3 as the feature extractor for WSOL methods to indicate our generalization for the backbone other than ResNet. The results are in accordance with utilizing ResNet, improving the performance on all baseline methods, for example, 11.31\% and 9.38\% G-Loc improvement for the vanilla structure (CAM) and DAOL on the CUB-200 dataset, respectively. Moreover, our BagCAMs still outperforms other SOTA methods with InceptionV3 on nearly all metric for these three datasets.

Localization maps generated by WSOL methods with our BagCAMs are also visualized in Fig.~\ref{fig:results}. For localization maps of the vanilla structure, only the most discriminative locations are activated, \textit{e.g.}, the pedestal of the toy, both ends of the pillar, the shade of the lamp, and the head of the bird. Though existing WSOL methods catch more positions of objects, they only enlarge or refine the activation of regions that near the discriminative parts rather than catching more parts of the object. This also visually verifies that the mechanism of CAM limits the performance of these WSOL methods, making the localizer only concern global cues. Profited by the utilization of our base localizer set, more object parts are effectively activated when adopting our BagCAMs to replace CAM for those methods, for example, the head of the toy, the pedestal of the lamp, and the body of the pillar/bird. Moreover, our BagCAMs can generate the localization map on intermediate layers that contains more fining cues such as pixels near the edge of objects, which also contributes to our high performance. 

\begin{minipage}{\textwidth}
 \begin{minipage}[t]{0.5\textwidth}
  \centering
     \makeatletter\def\@captype{table}\makeatother
     \caption{The best scores of different CAMs on layers of ResNet for CUB-200 dataset}
     \setlength{\tabcolsep}{0.5pt}
\begin{tabular}{c|ccccc}
\toprule
 & T-Loc & G-Loc & B-Loc & pIoU & PxAP \\
\hline
CAM
& 55.31 & 66.06 & 59.21 & 46.70 & 65.94
\\
PCS 
& 60.27& 73.93& 65.24& 52.05& 72.06
\\
Grad 
& 56.68 & 69.93 & 61.70 & 49.51 & 68.69
\\
Grad++ 
& 61.10& 73.79& 69.14& 53.61& 76.33
\\
\textbf{Ours} 
& \textbf{70.89} & \textbf{87.44} & \textbf{76.22} & \textbf{64.40} & \textbf{84.38}
\\
\hline
HAS  
& 54.48 & 72.55 & 66.25 & 51.00 & 71.87
\\
PCS 
& 53.65& 73.24& 67.9& 54.87& 76.72
\\
Grad
& 56.82& 77.79& 69.37& 55.64& 76.77
\\
Grad++
& 55.31 & 76.82& 70.29& 53.34& 75.54
\\
\textbf{Ours} 
& \textbf{65.93} & \textbf{89.65} & \textbf{84.45} & \textbf{70.24} & \textbf{88.94}
\\
\hline
CutMix 
& 56.27 & 64.13 & 59.08 & 44.21 & 65.23
\\
PCS 
&57.65 &68.13& 61.51& 48.19& 68.74
\\
Grad
&60.96& 72.68& 64.50& 52.10& 71.49
\\
Grad++
&63.17& 77.10& 67.67& 53.77& 74.78
\\
\textbf{Ours} 
& \textbf{72.96} & \textbf{87.44} & \textbf{79.67} & \textbf{64.93} & \textbf{85.36}
 \\
\hline
ADL  
& 52.13 & 66.75& 59.31& 45.40 & 59.49
\\
PCS 
& 52.13 & 66.75& 59.31& 45.40& 59.49
\\
Grad
& 52.13 & 66.75& 59.31& 45.40& 59.49
\\
Grad++
& 53.65& 70.89& 61.19& 44.72& 61.14
\\
\textbf{Ours} 
& \textbf{64.41} & \textbf{86.06} & \textbf{74.48} & \textbf{60.46} & \textbf{81.07}
\\
\hline
DAOL
& 62.40 & 81.83 & 69.87 & 56.18 & 74.70
\\
PCS 
& 63.30 & 84.57 & 71.49 & 58.94 & 76.81
\\
Grad
& 63.30 & 84.57 & 71.49 & 58.94 & 76.81
\\
Grad++ 
& 66.13 & 89.60 & 75.71& 63.08 & 80.23
\\
\textbf{Ours} 
& \textbf{69.67} & \textbf{94.01} & \textbf{84.88} & \textbf{74.51} & \textbf{90.38}
\\
\bottomrule
\end{tabular}
\label{tab:cams}
  \end{minipage}
  \begin{minipage}[t]{0.5\textwidth}
   \centering
\makeatletter\def\@captype{table}\makeatother
\caption{PxAP on layers of ResNet}
\setlength{\tabcolsep}{1pt}
\centering	
\begin{tabular}{c|ccccccc}
\toprule
Method & $Res_{1}$ & $Res_{2}$ & $Res_{3}$ & $Res_{4}$ \\
\hline
PCS 
& 42.01& 51.36& 72.96& 65.94
\\
Grad
& 15.05 & 19.61 & 68.69 & 65.94 
\\
Grad++ 
& 13.16& 32.01& 76.33& 71.49
\\
\textbf{Ours} 
& \textbf{71.35} & \textbf{78.71} & \textbf{84.38} & \textbf{72.98}
\\
\bottomrule
\end{tabular}
\label{tab:layer_res}
\makeatletter\def\@captype{table}\makeatother
\caption{PxAP on layers of Inception}
\centering	
\setlength{\tabcolsep}{0.5pt}
\begin{tabular}{c|ccccc}
\toprule
Method & $Mix_{6b}$ & $Mix_{6c}$ & $Mix_{6d}$ & $Mix_{6e}$ \\
\hline
PCS 
& 41.42& 61.37& 75.36& 76.32
\\
Grad
& 28.91 & 46.62 & 65.41 & 76.19 
\\
Grad++ 
& 26.77& 43.26& 65.41& 68.00
\\
\textbf{Ours} 
& \textbf{78.14} & \textbf{81.46} & \textbf{82.80} & \textbf{81.49} 
\\
\bottomrule
\end{tabular}
\label{tab:layer_inception}
\makeatletter\def\@captype{table}\makeatother
         \caption{Efficiency (fps) of CAMs}
\centering	
\setlength{\tabcolsep}{0.5pt}
\begin{tabular}{c|ccccccc}
\toprule
Method & $Res_{1}$ & $Res_{2}$ & $Res_{3}$ & $Res_{4}$ \\
\hline
PCS 
& \textbf{90.88} & 90.40 & \textbf{91.86} & \textbf{91.04}
\\
Grad
& 89.72 & \textbf{91.04} & 90.94 & 90.75 
\\
Grad++ 
& 90.61& 89.25 & 90.62& 89.67
\\
\textbf{Ours} 
& 88.44 & 86.40 & 87.02 & 87.77
\\
\bottomrule
\end{tabular}
\label{tab:fps}
\makeatletter\def\@captype{table}\makeatother\caption{Different Weight Strategy}
\centering	
\setlength{\tabcolsep}{0.5pt}
\begin{tabular}{c|cccccc}
\toprule
 & T-Loc & G-Loc & B-Loc & PxAP \\
\hline
CAM
& 55.31 & 66.06 & 59.21 & 65.94
\\
Ours$_{1}$
& 66.75 & 82.34 & 74.80 & 77.29
\\
Ours$_{2}$ 
& 70.20& 86.20& 74.16 & 81.79
\\
Ours$_{3}$ 
& 70.89 & 87.44 & 76.22 & 84.38
\\
\bottomrule
\end{tabular}
\label{tab:ablation}
   \end{minipage}
\end{minipage}

\begin{figure}
\centering
\includegraphics[width=1.00\textwidth]{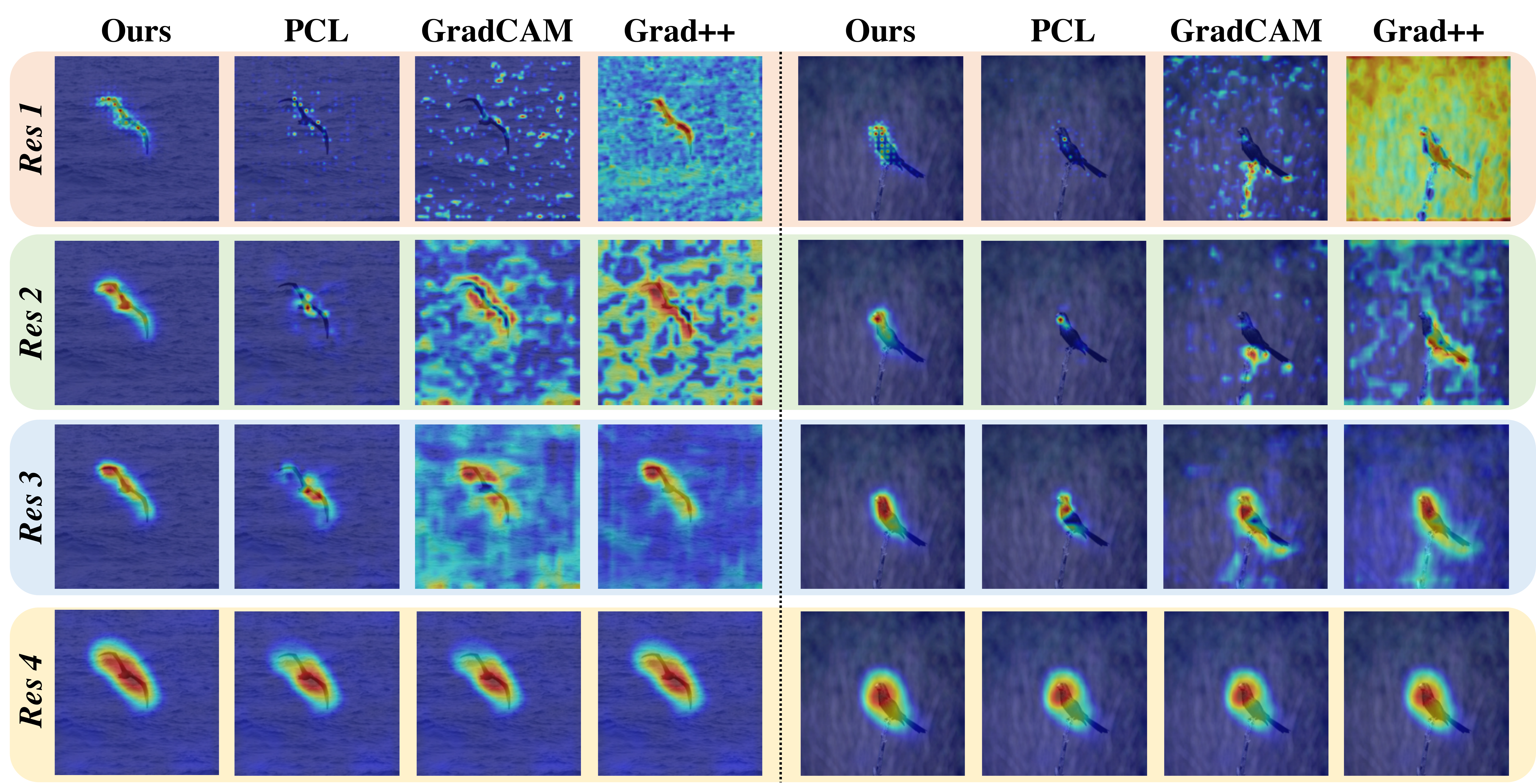}
\caption{Localization map generated by the features of different ResNet layer by CAMs.}
\label{fig:res_layer}
\end{figure}

\subsection{Discussion}

To deeply investigate the effectiveness of our BagCAMs, we also conducted  experiments to compare it with methods that generalize or enhance CAM on GAP-free structures or intermediate layers of CNN, \textit{e.g.}, GradCAM (Grad)~\cite{GradCAM}, GradCAM++ (Grad++)~\cite{GradCAM++}, PCS~\cite{CEM}. We adopted the same trained checkpoint for these methods and utilized them to project the classifier in the test process. Except for the original CAM, other methods can be added onto the intermediate layers of the feature extractor. Thus, we generated localization maps on each layer and chose the best performance to report.

Corresponding results are illustrated in Table~\ref{tab:cams} where the baseline methods, \textit{i.e.}, CAM, HAS, ADL, CutMix, and DAOL, represent directly adopting the classifier for localization as CAM. It shows that for all baseline WSOL methods, our BagCAMs achieves the highest improvement compared with other CAM mechanisms. This is because other CAMs methods all initialize $\hat{\bm{P}}$ with the global classification results $\textbf{s}$ for all positions as discussed in Sec.~\ref{sec:genrealize}, resulting in their lower improvement. Unlike them, our BagCAMs adopts $\bm{\hat{P}}_{:, m}$ to distribute a specific initial localization score for each position $m$, which helps generate valid localizers and contributes to our outstanding improvement, \textit{e.g.}, 15.68\% higher PxAP than using the original CAM for the DAOL. 

In addition, our BagCAMs can also achieve satisfactory performance when localizing objects based on features of intermediate layers, which may inspire generating localization maps with higher resolution to consider more details. Table~\ref{tab:layer_res} illustrates the PxAP metric for generating localization maps based on the feature of $Res_1$ ($256\times56\times56$), $Res_2$ ($512\times28\times28$), $Res_3$ ($1024\times28\times28$), and $Res_4$ ($2048\times28\times28$). Note that the original CAM can only adopt to the last layer before GAP due to the difference between the number of channels in $\mathbf{W}$ and the intermediate features, thus we did not include the original CAM in this Table~\ref{tab:layer_res}. It can be seen that GradCAM and GradCAM++ have great performance drops when projected to the prior intermediate layers, \textit{i.e.}, $Res_1$ and $Res_2$. Though the PCS, proposed for generating localization results on intermediate layers, slightly decelerates this decline, its PxAP of $Res_1$ is still $30.97\%$ lower than $Res_4$. Compared with them, our BagCAMs generates the localization map by bagging the performance of $N \times N$ base localizers, where $N$ is the spatial resolution of the feature map. Thus, for the previous layers with higher resolution, more basic localizers can be defined for bagging, \textit{i.e.}, $3,136$ for $Res_{1}$. This makes our BagCAMs achieve 29.34\% higher PxAP compared with the best of others, when projected on the feature of $Res_{1}$.

Fig.~\ref{fig:res_layer} also qualitatively visualizes the localization maps generated on the intermediate features. It can be seen that localization maps of GradCAM and GradCAM++ contain more noise on $Res_1$ and $Res_2$, and the PCS only activates a few discriminative locations. Compared with them, though our BagCAMs suffers from the grid effect caused by the down-sampling, our localization map can cover more object parts even for $Res_1$. Finally, the efficiencies of different CAMs are also shown in Table.~\ref{tab:fps}, where their mean frame per second (fps) for inferring CUB-200 test are reported. It can be seen that, though considering multiple regional localizers rather than only the global one, the complexity of our BagCAMs is only a bit higher than other methods. This indicates that our method can balance the localization performance and efficiency well.

Except for comparing with other CAM mechanisms, the choice of different weighting strategies, \textit{i.e.}, various settings of the weight matrix $\mathbf{\Lambda}$, were also explored on the CUB-200 dataset. Specifically, we designed three types of BagCAMs: (1) Ours$_1$ that only averages the scores generated by localizers $f^{m}_{n}$, \textit{i.e.}, $\mathbf{\Lambda}^{i}=\frac{1}{N}\mathbf{I}$. (2) Ours$_2$ that aggregates the scores with the spatially weighting mechanism of GradCAM++~\cite{GradCAM++}, \textit{i.e.}, $\mathbf{\Lambda}^{i}=diag({\alpha})$. (3) Ours$_3$, the mechanism we used in our paper as defined in Eq.~\ref{eq:setting}, which only selects specific localizers for each position like PCS~\cite{CEM}. Corresponding results are shown in Table~\ref{tab:ablation}. It can be seen that using these three weighting mechanisms can all enhance the performance of the baseline methods, profited by adopting the regional localizer set rather than the globally defined classifier. Specifically, when simply averaging the localization scores of the regional localizers (Ours$_1$), the performance improves about $11.35\%$ on the PxAP metric. Adopting the spatial weighting strategy to consider the effect based on each spatial position will bring an additional $4.5\%$ improvement. When grouping the $N*N$ localizers into $N$ clusters that are specifically used for each spatial position to reduce noise as PCS~\cite{CEM}, the performance hits the highest, \textit{i.e.}, about 84.38\% PxAP. Thus, we suggest adopting this grouping strategy to weight the effect of the regional localizers.

\subsection{Conclusion}

This paper proposes a novel mechanism called BagCAMs for WSOL to replace CAM~\cite{CAM} when projecting an image-level trained classifier as the localizer to locate objects. Our BagCAMs can be engaged in existing WSOL methods to improve their performance without re-training the baseline structure. Experiments show that our method achieves SOTA performance on three WSOL benchmarks. 

\subsubsection{Acknowledgements} This work was supported in part by the Beijing Natural Science Foundation under Grant Z210008; in part by the Shenzhen Science and Technology Program under Grant KQTD20180412181221912 and Grant JCYJ20200109140603831.

%
%
\bibliographystyle{splncs04}
\bibliography{egbib}
\end{document}